\def\year{2020}\relax
\documentclass[letterpaper]{article} 
\usepackage{aaai20-arXiv}  
\usepackage{times}  
\usepackage{helvet} 
\usepackage{courier}  
\usepackage[hyphens]{url}  
\usepackage{graphicx} 
\usepackage{graphicx}  
\usepackage{amsmath,amssymb,amsfonts}
\usepackage{amsthm}
\usepackage{algorithmic}
\usepackage{textcomp}
\usepackage{bm}
\usepackage{subfigure}
\usepackage[boxed,ruled,lined]{algorithm2e}
\usepackage[american]{babel}
\usepackage{setspace}
\usepackage{enumerate}
\usepackage{multirow}
\usepackage{url}

\newtheorem{theorem}{Theorem}

\newtheorem{definition}{Definition}
\newtheorem{remark}{Remark}

\title{Variance Reduction for Deep Q-Learning using Stochastic Recursive Gradient}
\author{Haonan Jia,\textsuperscript{\rm 1} Xiao Zhang,\textsuperscript{\rm 1} Jun Xu,\textsuperscript{\rm 1}\thanks{Corresponding author.} Wei Zeng,\textsuperscript{\rm 2}
Hao Jiang,\textsuperscript{\rm 3} Xiaohui Yan,\textsuperscript{\rm 3} Ji-Rong Wen\textsuperscript{\rm 1}\\
\textsuperscript{\rm 1}Gaoling School of Artificial Intelligence, Renmin University of China, Beijing 100872, China\\ 
\textsuperscript{\rm 2}Baidu Inc. 
\textsuperscript{\rm 3}Huawei Technologies\\
junxu@ruc.edu.cn
}
 
\begin{document}

\maketitle

\begin{abstract}
Deep Q-learning algorithms often suffer from poor gradient estimations with an excessive variance, resulting in unstable training and poor sampling efficiency. Stochastic variance-reduced gradient methods such as SVRG have been applied to reduce the estimation variance~\cite{zhao:2019:stochastic-arXiv}. However, due to the online instance generation nature of reinforcement learning, directly applying SVRG to deep Q-learning is facing the problem of the inaccurate estimation of the anchor points, which dramatically limits the potentials of SVRG. To address this issue and inspired by the recursive gradient variance reduction algorithm SARAH~\cite{nguyen:2017:sarah}, this paper proposes to introduce the recursive framework for updating the stochastic gradient estimates in deep Q-learning, achieving a novel algorithm called SRG-DQN. Unlike the SVRG-based algorithms, SRG-DQN designs a recursive update of the stochastic gradient estimate. The parameter update is along an accumulated direction using the past stochastic gradient information, and therefore can get rid of the estimation of the full gradients as the anchors. Additionally, SRG-DQN involves the Adam process for further accelerating the training process. Theoretical analysis and the experimental results on well-known reinforcement learning tasks demonstrate the efficiency and effectiveness of the proposed SRG-DQN algorithm.
\end{abstract}

\section{Introduction}
Recent years have witnessed the dramatic progress of deep reinforcement learning (RL) in a variety of challenging tasks including computer games, robotics, natural language process, and information retrieval. Amongst the RL algorithms, deep Q-learning is a simple yet quite powerful algorithm for solving sequential decision problems \cite{Mnih:2013:DQN,Mnih:2015:DQN}. Roughly speaking, deep Q-learning makes use of a neural network (Q-network) to approximate the Q-value function in traditional Q-learning models. The system state is given as the input and the Q-values of all possible actions are generated as the output.
The learning of the parameters in Q-network amounts to sequences of optimization problems on the basis of the stored agent's experiences. Stochastic gradient descent (SGD) is often employed to solve these optimization problems. That is, at each iteration of the optimization, to calculate the parameter gradients, the agent samples an action according to the current Q-network, issues the action to the environment, gathers the reward, and moves to the next state. The reward is used as the supervision information for calculating the gradient for updating the Q-network parameters. The gradient points in the direction of maximum increase the possibility of getting high accumulative future rewards.

In real-world applications, the stochastic gradient descent in deep Q-learning often suffers from the inaccurate estimation of the gradients. In other words, the estimated stochastic gradient has a large variance as it is estimated on the basis of the data collected from only one step action or a small batch of actions. The high variance gradient inevitably hurts the efficiency and effectiveness of the deep Q-learning algorithms. How to reduce the variance has become one of the key problems in deep RL. Research efforts have been undertaken to solve this problem. For example, Averaged-DQN~\cite{anschel:2017:averaged} extends the traditional DQN algorithm by averaging the previously learned Q-values estimates, achieving a variance reduced gradient estimation with an approximation error guarantee. More recently, \citeauthor{zhao:2019:stochastic} (\citeyear{zhao:2019:stochastic})  proposed an optimization strategy by combining the stochastic variance reduced gradient (SVRG)~\cite{johnson:2013:accelerating} technique and the deep Q-learning, called SVR-DQN. 
More methods on variance reduction for deep Q-learning please refer to~\cite{romoff:2018:reward,Sabry2019Reduction}.
SVRG has also been applied to policy gradient methods in RL as an effective variance-reduced technique for stochastic optimization,
such as off-line control~\cite{Xu2017Stochastic},
policy evaluation~\cite{Du2017Stochastic},
and on-policy control~\cite{Papini2018Stochastic}.

Though preliminary successes have been achieved, current methods are far from optimal because they ignored the essential differences between reinforcement learning and traditional machine learning. The SVRG-based methods need to pre-calculate full gradients as the anchors. The anchors are crucial for finding more accurate gradient direction estimations in the down-stream parameter update. When being executed on a fixed training set (e.g., in traditional machine learning settings), SVRG-based methods can easily estimate the anchors by scanning all of the training instances. In deep Q-learning, however, the \emph{anchors cannot be accurately estimated anymore} because the learning is conducted in an online manner: (1) In deep Q-learning, the training instances (i.e., the sampled transitions) are gradually generated with the training goes on, via issuing actions to the environment at each iteration. The algorithm cannot access the instances that will be generated in the future iterations; (2) In deep Q-learning, the selection of the actions is guided by the DQN with current parameters. Therefore, the generated instances at different iterations cannot be identically distributed, as the DQN parameters have been updated. The phenomenon makes the problem of inaccurate estimation of the anchors more severe. Empirical analyses also have shown that the inaccurate estimation of the anchors greatly impacted the performances of the SVRG-based methods.

In this paper, to address the issue and inspired by the variance reduction algorithm SARAH~\cite{nguyen:2017:sarah}, we propose to adopt the recursive gradient estimation mechanism in SARAH into the training iterations of deep Q-learning, achieving a novel deep Q-learning algorithm called SRG-DQN. Specifically, SRG-DQN contains and outer loop which samples $N$ training instances (i.e., $N$ transitions including the state, action, reward, and next-state) based on the current Q-network and from the experience pool, and an inner loop which first estimates the stochastic gradients recursively and then updates the Q-network parameters. Besides, the Adam process is executed at the end of the outer loop for further improving the efficiency of the training.
Theoretical and experimental analyses demonstrate that the recursive gradient estimation mechanism successfully addresses the problem of inaccurately anchors in SVRG-based methods. It also heritages the advantages from SARAH including the fast convergence rate, and the stable and reliable training.
We conducted experiments on Atari Games to evaluate the proposed SRG-DQN algorithm. Experimental results show that SRG-DQN outperforms the state-of-the-art baselines including SGD-based and SVRG-based deep Q-learning algorithms, in terms of reward scores, convergence rates, and training time. Empirical analyses also show that SRG-DQN dramatically reduces the variance of the estimated gradients, discovering how and why SRG-DQN can improve the performance of the baseline algorithms.


\section{Background: Variance Reduction for Deep Q-Learning}
Reinforcement learning (RL) is a branch of machine learning wherein the agent learns to perform certain actions in an environment which leads it to maximize the cumulative future rewards. Q-learning is a type of widely used RL algorithms which learns the action-value function (Q-function) to choose the actions.
In the literature of deep Q-learning,
most of the research focuses on the improvement of algorithm components and model structures~\cite{van:2016:DDQN,schaul:2015:prioritized,wang:2015:dueling,fortunato:2017:noisy,hessel:2018:rainbow}.
In this paper,
we focus on the topic of the variance reduction for deep Q-learning.
\subsection{SGD for Deep Q-Learning}
In Q-learning, it is assumed that the agent will perform the sequence of actions that will eventually generate the maximum total reward (return). The return is also called the Q-value and the strategy is formalized as:
\[
Q(s, a) = r(s, a) + \gamma \max_{a\in \mathcal{A}(s)} Q(s', a),
\]
where $\gamma \in [0, 1]$ is a discount factor which controls the contribution of rewards in the future, $s$ is the current state, $\mathcal{A}(s)$ contains all of the candidate actions under state $s$, $a$ is the selected action, $r(s,a)$ is the received reward after issuing $a$ at $s$, and $s'$ is the next state that the system moves to after issuing $a$. The equation states that the Q-value yielded from being at state $s$ and performing action $a$ is the immediate reward $r(s,a)$ plus the highest Q-value possible from the next state $s'$. It is easy to see that $Q(s,a)$ helps the agent to figure out exactly which action to perform.

Traditionally, Q-value is defined as a table with which the agent figure out exactly which action to perform at which state. However, the time and space complexities become huge when facing large state and action spaces. Deep neural networks have been used to approximate the Q-value function, called deep Q-learning. The learning of the parameters in the Q-network amounts to a serious of optimization problems. Specifically, assuming that at time step $t$, the system is at state $S_t$ and the agent issues an action $A_t$. After that at the time step $t+1$, it receives a reward $R_{t+1}$ and transits to state $S_{t+1}$. Therefore, we collect a transition tuple $(S_t, A_t, R_{t+1}, S_{t+1})$. The loss function, therefore, is defined as the mean squared error of the Q-value predicted by the Q-network and the target Q-value, where the target Q-value is derived from the Bellman equation
$
y(S_t, A_t) = R_{t+1} + \gamma \max_{a\in\mathcal{A}(S_t)}Q(S_{t+1}, a;\mathbf{\theta}),
$
where $Q(S_{t+1}, a;\mathbf{\theta})$ is the Q-network with parameters $\mathbf{\theta}$ that predicts the Q-value for the next state with the same Q-network. Stochastic gradient descent has been employed for conducting the optimization. Given a sampled transition $(S, A, R, S')$, the stochastic gradient can be estimated as:
\begin{align*}
\mathbf{g}  &= \nabla (y(S, A) - Q(S, A; \theta))^2 \\
            &= 2 (y(S, A) - Q(S, A; \theta))\nabla Q(S, A;\theta),
\end{align*}
where $\nabla Q(S, A;\theta)$ calculates the gradient of $Q$ w.r.t. the parameter $\mathbf{\theta}$.

\subsection{Variance Reduced Deep Q-Learning}
The original stochastic gradient descent based on a single transition often hurts from the problem of high gradient estimation variances.
There are many ways to accelerate SGD convergence from the perspective of reducing variance,
such as SAG~\cite{roux:2012:sag}, SAGA~\cite{defazio:2014:saga}, and SVRG~\cite{johnson:2013:accelerating}.
Researchers have combined the variance reduction techniques proposed in traditional machine learning with deep Q-learning. For example,
\citeauthor{zhao:2019:stochastic-arXiv} (\citeyear{zhao:2019:stochastic-arXiv})   proposed an algorithm called Stochastic Variance Reduction for Deep Q-learning (SVR-DQN) which combines SVRG with utilizes the optimization strategy of SVRG during the learning. Specifically, at each outer iteration $s = 1, 2, \cdots, S$, the algorithm samples a batch of $N$ training transitions $\mathcal{D} = \{(S_i, A_i, R_{i+1}, S_{i+1})\}_{i=1}^{N}$, and calculates a full gradient on $\mathcal{D}$ as the anchor:
\[
\tilde{\mathbf{g}} = \frac{1}{N}\sum_{i=1}^N 2(y_i - Q(S_i, A_i;\mathbf{\theta}_0^s)) \nabla Q(S_i, A_i;\mathbf{\theta}_0^s),
\]
where
$
y_i = R_{i+1} + \gamma\max_{a\in\mathcal{A}(S_{k+1})}Q(S_{k+1}, a; \theta_0^s),
$
and $\mathbf{\theta}_0^s$ is the network parameter at the beginning of the $s$-th outer iteration.
In its inner iteration indexed by $m$, for each sampled transition $(S,A,R,S')\in \mathcal{D}$, the stochastic gradients w.r.t. up to date parameters are calculated:
\[
\mathbf{g}_m^s = 2(y_m - Q(S,A;\theta_m^s))\nabla Q(S,A;\theta_m^s),
\]
where
$
y_m = R + \gamma \max_{a'\in \mathcal{A}(S')} Q(S',a';\theta_m^s).
$
Similarly, the stochastic gradients w.r.t. `old parameters' are also calculated:
$$
\mathbf{g}_0^s = 2(y_0 - Q(S,A;\theta_0^s))\nabla Q(S,A;\theta_0^s),
$$
where
$$
    y_0 = R + \gamma\max_{a'\in \mathcal{A}(S')}  Q(S',a';\theta_0^s).
$$
Finally the variance reduced gradient is calculated as
$$
\Delta = \mathbf{g}^s_m - \mathbf{g}^s_0 +\tilde{\mathbf{g}}.
$$
Besides,
at each outer iteration,
SVR-DQN obtains a more accurate estimation of the gradient using Adam process,
which can accelerate the training of deep Q-learning and improve the performances~\cite{zhao:2019:stochastic-arXiv}.

\section{Our Approach: SRG-DQN}
In this section, we analyze the limitations of the variance reduction mechanism in SVR-DQN and propose a novel deep Q-learning algorithm called SRG-DQN.

\subsection{Problem Analysis}
In SVR-DQN, the accurate estimations of the anchors $\tilde{\mathbf{g}}$ are crucial they provide stable baselines to adjust the stochastic gradient. In traditional machine learning, the SVRG anchors can be accurately estimated based on the whole train data (i.e., full gradients). In deep Q-learning, however, the training instances are not fixed in advance but we need to collect them at each parameter change. Therefore, the estimated anchors are only based on $N$ instances sampled at the previous and current iterations.  


The phenomenon inevitably makes the estimated anchors inaccurate due to the following two reasons.
First, deep Q-learning algorithms are usually run in an online manner by nature. At each iteration, the algorithm samples an (or some) instance(s) as the new training data. Therefore, it is impossible for the algorithm to estimate the full gradients as the anchors, because only a part of the whole training instances (the instances sampled at the past iterations) are accessible.
Second, the instances sampled at different iterations are based on the Q-network with different parameters. That is, at each iteration, the agent first sample an action guided by the Q-network $Q(s,a;\theta)$ (for example, $\epsilon$-greedy), and then use the sampled instance to update network parameters. Therefore, the Q-network is continuously updated during the training, and the sampled instances at different iterations would belong to different distributions. This will make the estimated anchors cannot reflect the directions that the exact gradients should point to.
\begin{figure}[htb!]
    \centering
    \includegraphics[width=0.45\textwidth]{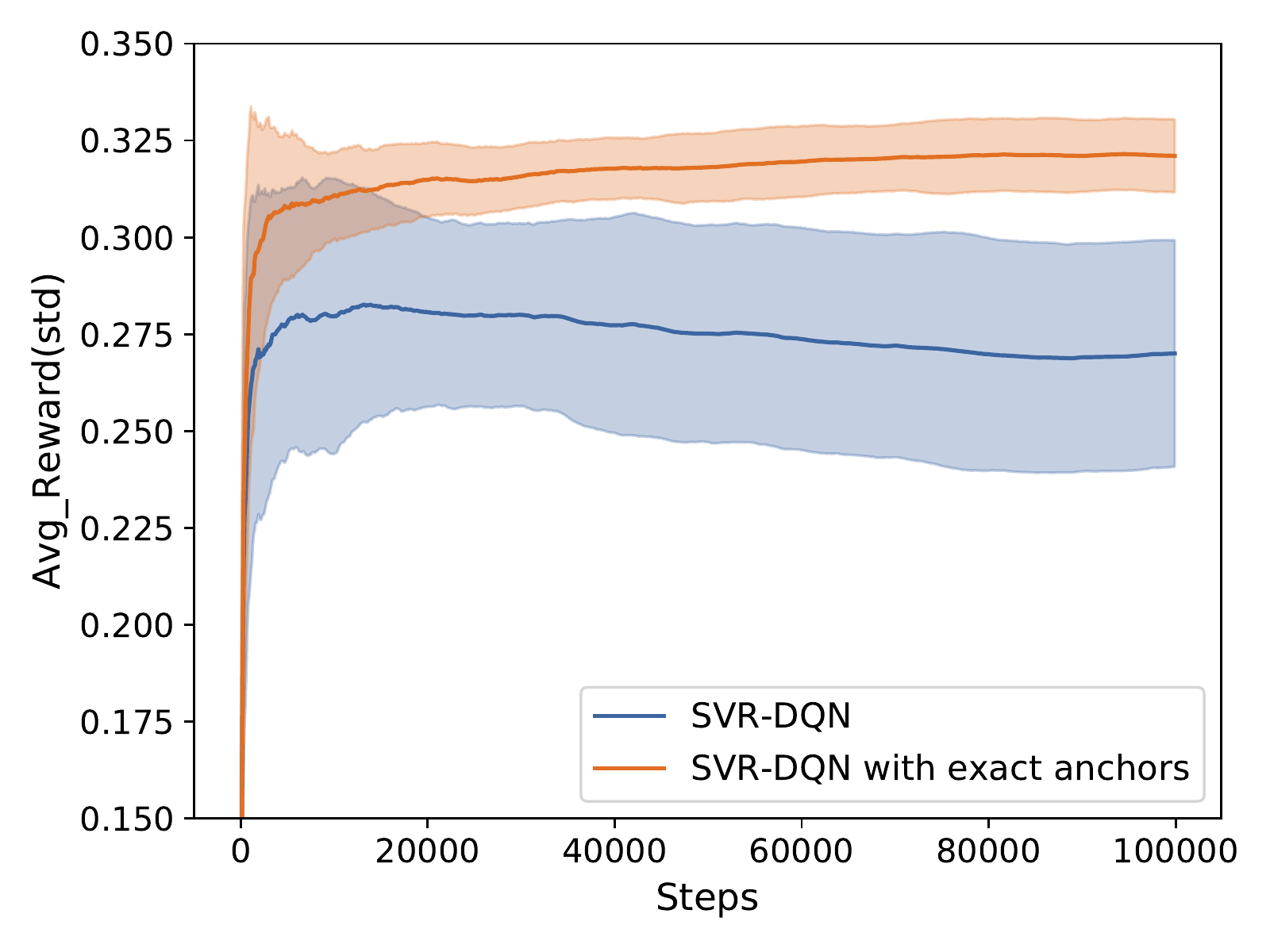}
    \caption{Training curves of ``SVR-DQN'' and ``SVR-DQN with exact anchors''. The shaded area presents one standard deviation. We focus on the influence of anchors and omit the Adam process in these algorithms.}
    \label{fig:analysis}
\end{figure}

We conducted an experiment to show the impact of the inaccurate anchors. Specifically, based on the OpenAI gym Mountain Car, we compared the performances of deep Q-learning with SVRG (SVR-DQN) and its variation in which the anchor points \emph{could be exactly estimated}. To do this, we first ran the existing SVR-DQN and recorded all of the sampled transitions. Then, we re-ran SVR-DQN using identical settings and training transitions at each iteration, except estimating the $\tilde{\mathbf{g}}$ in line 5 with all of the recorded transitions (denoted as ``SVR-DQN with exact anchors''). In this way, the anchors $\tilde{\mathbf{g}}$ are estimated on the whole training data as that of in traditional machine learning\footnote{Please note that ``SVR-DQN with exact anchors'' is not feasible in the real-world because the RL algorithms don't have the oracle view to access the instances sampled in the future iterations.}. Figure~\ref{fig:analysis} shows the training curves of the two models''. We can see that compared with ``SVR-DQN'',  ``SVR-DQN
with exact anchors'' converged faster, better, and with lower variances. The results clearly indicate that the inaccurate estimation of the anchors can hurt the power of SVRG. We conclude that directly applying SVRG to deep Q-learning violates the basic assumptions of SVRG, and therefore limits its full potentials.

\subsection{Recursive Gradient Deep Q-Learning}
To address the problem and inspired by the algorithm of SARAH~\cite{nguyen:2017:sarah}, in this paper we propose a novel algorithm called Stochastic Recursive Gradient Deep Q-Network (SRG-DQN). Different from the SVRG-based methods, SRG-DQN resorts to the recursive gradients rather than the full gradients, as the anchors. In this way, SRG-DQN gets rid of the inaccurate estimation of the anchors.

As shown in Algorithm~\ref{alg:SRG-DQN}, SRG-DQN contains an outer loop indexed by $s$. At each outer loop, $N$ training instances $\mathcal{D}$ are sampled and the initial anchor point $\Delta_0^s$ is calculated based on all of the $N$ sampled instances\footnote{Note that here $\Delta_0^s = \tilde{\mathbf{g}}$ in SVR-DQN.}:
\begin{align*}
\Delta_0^s  &= \nabla\left( \frac{1}{N}\sum_{i=1}^N (y_i - Q(S_i, A_i;\mathbf{\theta}_0^s))^2\right)\\
&=  \frac{1}{N}\sum_{i=1}^N 2(y_i - Q(S_i, A_i;\mathbf{\theta}_0^s)) \nabla Q(S_i, A_i;\mathbf{\theta}_0^s),
\end{align*}
where each of the targets
$$
    y_i = R_{i+1} + \gamma
    \max_{a\in\mathcal{A}(S_{i+1})}Q(S_{i+1}, a; \theta_0^s)
$$
is the Q-value derived from the Bellman equation, for $i=1,\cdots, N$. $\Delta_0^s$ is first used to update the parameters.

At each iteration of its inner loop $m$, the algorithm first randomly samples one training instance $(S, A, R, S')\in\mathcal{D}$. The stochastic gradient w.r.t. current up-to-date parameters, denoted as ${\mathbf{\theta}}_m^s$, is calculated as follows:
\begin{align*}
    \mathbf{g}_m^s  &= \nabla(y_m - Q\left(S,A;\theta_m^s)\right)^2\\
                &= 2\left(y_m - Q(S,A;\theta_m^s)\right)\nabla Q(S,A;\theta_m^s),
\end{align*}
where the target
$$
    y_m = R + \gamma \max_{a'\in \mathcal{A}(S')} Q(S',a';\theta_m^s).
$$
Similarly, the stochastic gradient w.r.t. the previous inner loop parameter $\mathbf{\theta}_{m-1}^s$ is also calculated:
\[
    \mathbf{g}_{m-1}^s =
    2\left(y_{m-1} - Q(S,A;\theta_{m-1}^s) \right)
    \nabla Q(S,A;\theta_{m-1}^s),
\]
where
$$
    y_{m-1} = R + \gamma \max_{a'\in \mathcal{A}(S')} Q(S',a';\theta_{m-1}^s).
$$

Following the recursive gradient defined in \cite{nguyen:2017:sarah}, the final gradient at current loop, $\Delta_m^s$, can be defined recursively. That is, using the previous loop gradient $\Delta_{m-1}^s$ as the anchor point to estimate the current loop gradient:
\[
\Delta_m^s = \mathbf{g}_m^s - \mathbf{g}_{m-1}^s + \Delta_{m-1}^s.
\]
Note that the anchor for the first loop (i.e., $\Delta_0^s, m = 0$) is the full gradient calculated in the outer loop.
To further improve the performances and in light of Adam optimizer~\cite{Kingma2014Adam}, we also propose to introduce the Adam process in SRG-DQN. Specifically, after the ending of each inner loop, an Adam process is executed for further updating the parameters, including calculating a biased first moment, a biased second raw moment, a bias-corrected first moment, and a bias-corrected second raw moment, and finally conducting the parameter updating. Lines 16-21 conduct the Adam process.
Following the practices in \cite{nguyen:2017:sarah,Li2019Convergence},
SRG-DQN takes $\theta^S_{M+1}$ as its final output.

\begin{algorithm}[tbh!]
	\caption{Stochastic Recursive Gradient for Deep Q-Learning (SRG-DQN)}
	\label{alg:SRG-DQN}
	\begin{algorithmic}[1]
	\REQUIRE Deep Q-function $Q$, \# epochs $S$, epoch size $M$, discount factor $\gamma$, step size $\eta$, Adam stepsize $\alpha$,  Adam parameters $\beta_1, \beta_2$
	\ENSURE Model parameters $\mathbf{\theta}$
	\STATE Initialize $\mathbf{\theta}^0_{M+1} \leftarrow \theta_0$
	\FOR{$s=1$ \TO $S$}
	    \STATE $\mathbf{\theta}_0^s \leftarrow \mathbf{\theta}_{M+1}^{s-1}$
	    \STATE sample $N$ transitions $\mathcal{D} = \{(S_i, A_i, R_{i+1}, S_{i+1})\}_{i=1}^{N}$ according to $Q(s,a;\mathbf{\theta}_{0}^{s})$
	    \STATE $\Delta_0^s\leftarrow \frac{1}{N}\sum_{i=1}^N 2(y_i - Q(S_i, A_i;\mathbf{\theta}_0^s)) \nabla Q(S_i, A_i;\mathbf{\theta}_0^s)$ \\
where $y_i = R_{i+1} + \gamma\max \limits_{a\in\mathcal{A}(S_{k+1})}Q(S_{k+1}, a; \theta_0^s)$
	    \STATE $\theta_1^s \gets \theta_0^s -\eta \Delta_0^s$\COMMENT{update with full gradient}
	    \FOR{$m=1$ \TO $M$}
	        \STATE randomly select a transition $(S,A,R,S')\in \mathcal{D}$
	        \STATE $y_m \gets R + \gamma \max \limits_{a'\in \mathcal{A}(S')} Q(S',a';\theta_m^s)$
            \STATE $\mathbf{g}_m^s \gets 2(y_m - Q(S,A;\theta_m^s))\nabla Q(S,A;\theta_m^s)$ \COMMENT{gradient w.r.t. up-to-date parameters}

            \STATE $y_{m-1} \gets R + \gamma \max \limits_{a'\in \mathcal{A}(S')} Q(S',a';\theta_{m-1}^s)$
            \STATE \mbox{$\mathbf{g}_{m-1}^s \gets 2(y_{m-1} - Q(S,A;\theta_{m-1}^s))\nabla Q(S,A;\theta_{m-1}^s)$} \COMMENT{gradient w.r.t. previous-iteration parameters}
            \STATE $\Delta_m^s \gets \mathbf{g}_m^s -\mathbf{g}_{m-1}^s + \Delta_{m-1}^s$\COMMENT{recursive gradient which using the previous one as the anchor}

            \STATE $\theta_{m+1}^{s} \gets \theta_{m}^{s} - \eta \Delta_m^s$ \COMMENT{update parameters}
	    \ENDFOR
	
	    \COMMENT{start Adam process}
	    \STATE $\mathbf{g}_s \gets \mathbf{g}_M^s$
	    \STATE $\mathbf{m}_s \gets \beta_1 \cdot \mathbf{m}_{s-1} + (1-\beta_1) \cdot \mathbf{g}_s$

       \COMMENT{update biased first moment estimate}
	    \STATE $\mathbf{v}_s \gets \beta_2 \cdot \mathbf{v}_{s-1} + (1-\beta_2) \cdot (\mathbf{g}_s\circ \mathbf{g}_s)$ \COMMENT{update biased second raw moment estimate, where `$\circ$' denotes element-wise product}
	    \STATE $\mathbf{\widehat{m}}_s \gets \mathbf{m}_s / (1-\beta_1^s) $\COMMENT{compute bias-corrected first moment estimate}
	    \STATE $\widehat{\mathbf{v}}_s \gets \mathbf{v}_s / (1-\beta_2^s) $\COMMENT{compute bias-corrected second raw moment estimate}
	    \STATE $\theta_{M+1}^{s} \gets \theta_M^{s} - \alpha \cdot\widehat{\mathbf{m}}_s / \left( \sqrt{\|\widehat{\mathbf{v}}_s\|} + \epsilon \right)$\COMMENT{update parameters}
	\ENDFOR
	\RETURN $\mathbf{\theta}_{M+1}^{S}$
	\end{algorithmic}
\end{algorithm}

\subsection{Theoretical Analysis}
We analyze the convergence of SRG-DQN as follows. In the inner loop, the optimization in DQN can be formulated as the empirical risk minimization problem:
$$ \min_{\theta \in \Theta} F (\theta) := \frac{1}{M}\sum_{i=1}^M f_i (\theta),$$
where
$$
    f_i (\theta) = \left[ y_i - Q(S_i, A_i; \mathbf{\theta}) \right]^2.
$$
The optimization problem in DQN is nonconvex due to the composite structure of neural networks.
Given the error parameter $\varepsilon > 0,$
the goal is to search for an \emph{$\varepsilon$-optimal point} $\theta \in \Theta$ such that
$$\mathbb{E} [ \| \nabla F(\theta) \|^2 ]\leq \varepsilon.$$

First, similar to that of in \cite{Agarwal2015Lower}, we give the definition of the Incremental First-order Oracle (IFO):
\begin{definition}
   IFO takes a point $\theta \in \Theta$ and an index $i \in \{1,2, \ldots, M\}$ as inputs
   and returns the pair $ \nabla f_i (\theta)$.
\end{definition}
Then the convergence rate of the algorithms can be measured by the oracle complexity.
The \emph{oracle complexity} is defined as the smallest number of queries to IFO leading to an $\varepsilon$-optimal point.
Further assuming that each function $f_i (\theta)$ is $\beta_i$-smooth and bounded for $i \in [M]$ (that is, $\nabla f_i (\theta)$ is $\beta_i$-Lipschitz continuous and $|f_i (\theta)| \leq B_i, ~i \in [M], \theta \in \Theta$), we have the following Theorem~\ref{thm:SRG-DQN:IFO:single_outer} for SRG-DQN:
\begin{theorem}\label{thm:SRG-DQN:IFO:single_outer}
    Let $\mu = \sum_{i=1}^{M} \beta_i^2 / M$ and $B_{\max} = \sup_{i \in [M]} \{B_i\}.$
    For our SRG-DQN 
    within a single outer loop (in outer iteration $s \in [S]$),
    setting $\eta \leq 2/ [\sqrt{\mu} ( \sqrt{4M +1} +1 ) ]$
    to attain an $\varepsilon$-optimal point requires
    $$
        \Omega \left(  \sqrt{M} / \varepsilon \right)
    $$
    queries to IFO.
\end{theorem}
Proof of Theorem~\ref{thm:SRG-DQN:IFO:single_outer} can be found in Appendix A.


\begin{remark}
    While the oracle complexity of SVRG is $$\widetilde{O}\left( M + M^{2/3} / \varepsilon \right)$$ for the optimization problem in DQN \cite{Li2019Convergence},
    our SRG-DQN achieves a lower oracle complexity w.r.t. the number of the epoch size $M$,
    which indicates SRG-DQN has a faster convergence rate than that of SVRG for DQN.
\end{remark}

\section{Experiments}

\subsection{Experimental Settings}
The experiments were conducted on the tasks `MountainCar-v0', `Pendulum-v0', and `cartpole-v1' of the OpenAI gym library. In the experiments, the deep $Q$-networks were set as the three-layer fully connected networks. Following the setups in~\cite{Mnih:2015:DQN}, $\epsilon$-greedy strategy was used for the action selection and the model exploration, where $\epsilon$ decreases linearly from initial value $0.1$ to $0.001$. 
The transfer instances generated during the interactions between the agent and the environment were stored in the experience replay memory,
which adopted a first-in-first-out mechanism to store the transition data.
When performing the gradient descent,
the algorithm sampled $64$ transition instances from the experience replay memory uniformly as the training batch data.
The learning frequency was set to $16$, which means that the batch data was sampled once every $16$ rounds.
In all the experiments, the DQN algorithm in~\cite{Mnih:2015:DQN} was adopted as the main process.
Our algorithms were called  ``SRG-DQN'' and ``SRG-DQN without Adam Process''.
The DQN optimized by SGD (called ``DQN with SGD'') and DQN optimized by SVRG (called ``SVR-DQN'')
were chosen as the baselines.
The detailed specification of the experimental setups was provided in Table~\ref{table:SRG_DQN:setups}.
\begin{table*}[htb!]
    \centering
    \footnotesize
    \caption{The detailed specification of experimental setups in our SRG-DQN.}
    \label{table:SRG_DQN:setups}
    \vskip 0.1in
    \begin{spacing}{1.2}
    \begin{tabular}{|l|r|r|r|r|}
      \hline
      Item & Notation & CartPole & MountainCar & Pendulum \\
      \hline
      \#~neural networks hidden nodes &-- & 8 & 20 & 20 \\
      neural networks activate function &-- & ReLU & ReLU & ReLU \\
      Adam stepsize & $\alpha$ & $10^{-3}$ & $10^{-3}$ & $10^{-3}$ \\
      step size & $\eta$ & $10^{-2}$ & $10^{-2}$ & $10^{-3}$ \\
      Adam parameter &$\beta_1$ &0.9 & 0.9 & 0.9\\
      Adam parameter &$\beta_2$ & 0.999 & 0.999 & 0.999\\
      batch size of training transitions &$N$  & 64 & 64 & 32\\
      number of inner iterations &$M$ & 16& 16& 16\\
      discount factor &$\gamma$& 0.99 & 0.9 & 0.9\\
      total number of episodes/steps & -- &800 episodes	&100,000 steps	&20,000 steps\\
      \hline
    \end{tabular}
    \end{spacing}
    \vskip 0.1in
\end{table*}

\begin{figure*}[htb!]
       \includegraphics[width=0.33\textwidth]{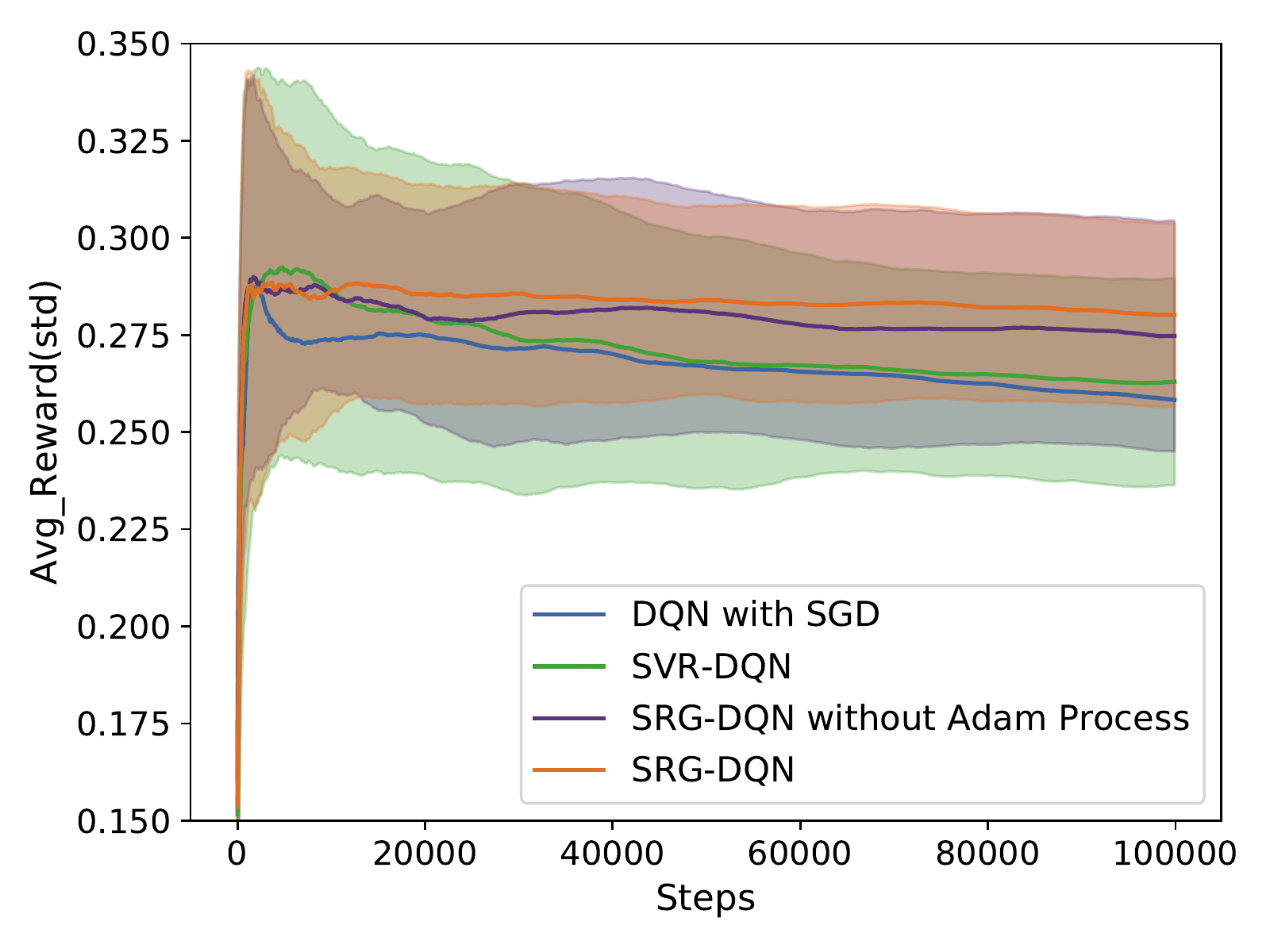}
        \includegraphics[width=0.33\textwidth]{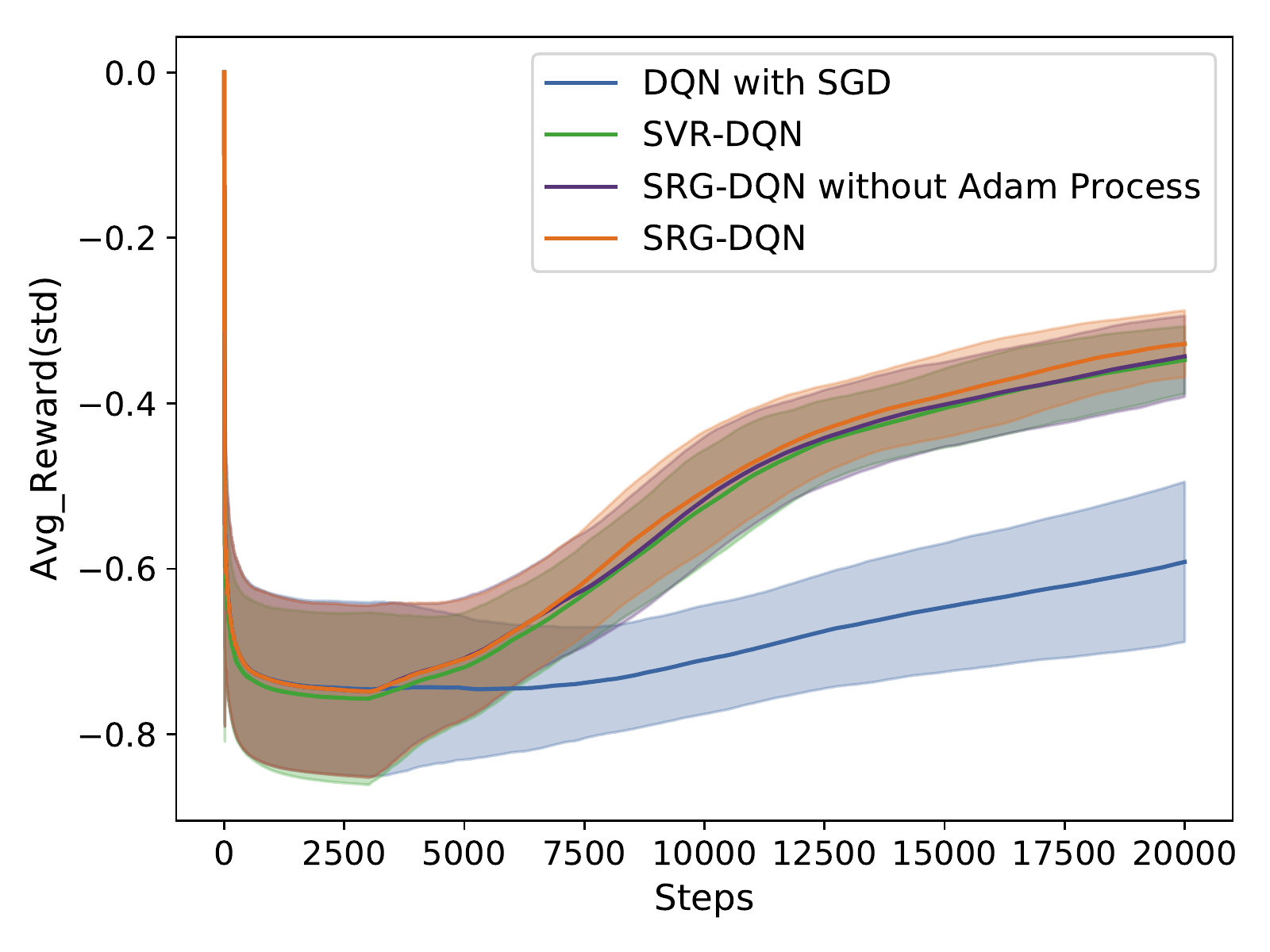}
        \includegraphics[width=0.33\textwidth]{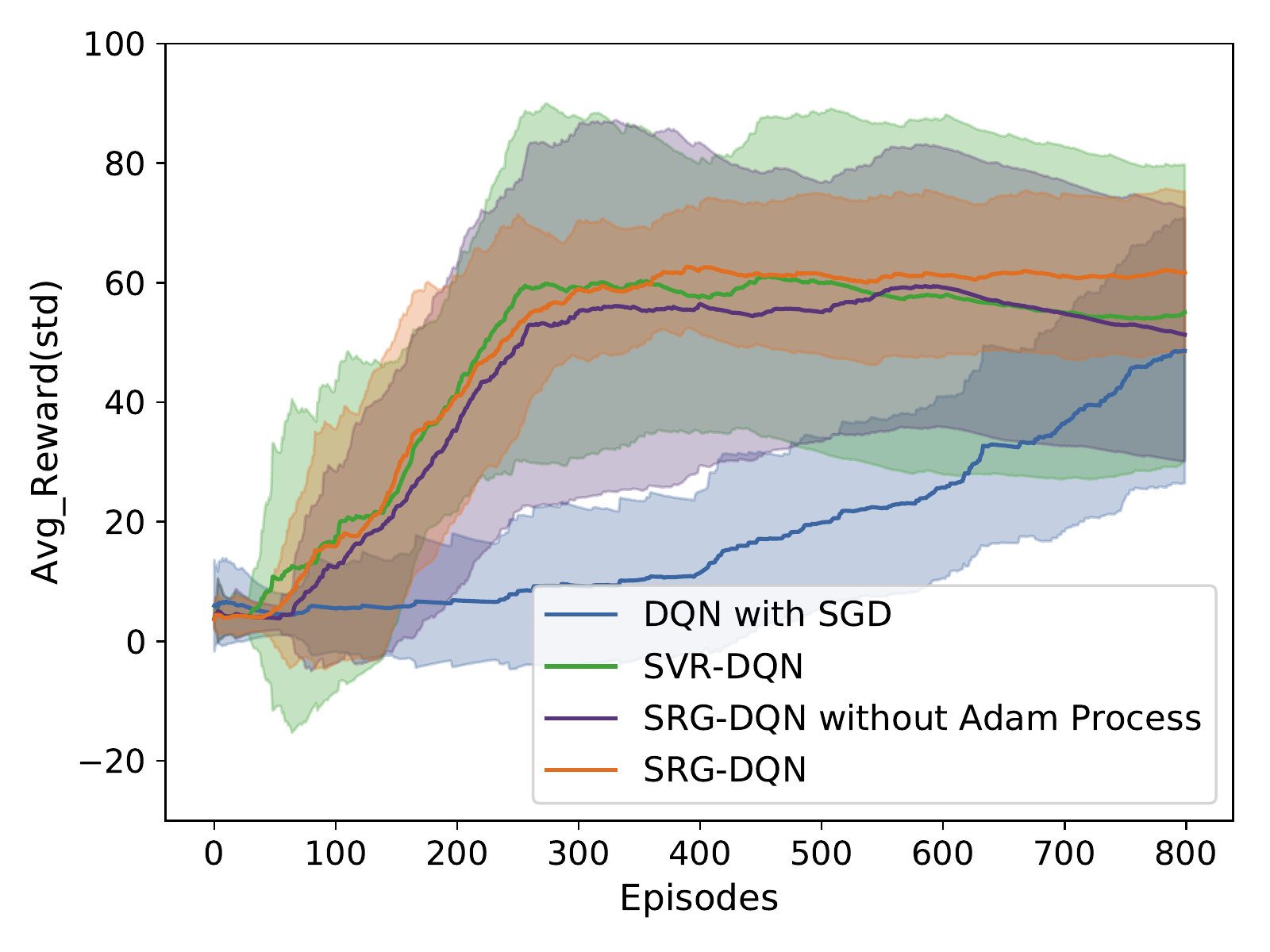}
    \caption{Performance curves of DQN with SGD (blue), SVR-DQN ~\cite{zhao:2019:stochastic-arXiv} (green),
    our SRG-DQN without Adam process (purple) and SRG-DQN (orange) for three tasks,
    where the shaded area represents a standard deviation.
    \emph{Left}: average rewards w.r.t. steps on `MountainCar-v0' task,
    where the `Steps' axis represents the outer iteration $s$ in Algorithm~\ref{alg:SRG-DQN} and the `Avg\_Reward' axis represents the average return of each action;
    \emph{Milddle}: average rewards w.r.t. steps on `Pendulum-v0' task,
    where the meanings of axes are the same as the left part in this figure;
    \emph{Right}: average rewards w.r.t. episodes on `cartpole-v1' task
    where the `Episode' axis represents a complete trajectory
    and
    the `Avg\_Reward' represents the average return per trajectory.
    }\label{fig:exp:perfCurves}
    \vskip 0.1in
\end{figure*}

\subsection{Experimental Results}
We conducted Avg\_reward experiments in three tasks,
in which the average reward is used to measure the performance of the agent.
For fair comparisons, DQN structures in all algorithms were set with the same parameters.

The left part of Figure ~\ref{fig:exp:perfCurves} compares the average performance of the four algorithms on the MountainCar task over 50 rounds. In this task, to encourage the car to explore, we replace the reward function from the original discrete value to a continuous function that is positively correlated with the car's altitude.
Without limiting the number of episodes, the four algorithms all run 100,000 steps. This means that the faster the car reaches the goal, the higher the average reward of each action. From the results, our algorithm can not only maintain a higher average reward, but also have better stability and lower variance after convergence.
We omit the standard deviation of DQN with SGD in this figure, which is obviously underperforming.
The middle part of Figure ~\ref{fig:exp:perfCurves} compares the performance of the four algorithms under Pendulum task.
In this task, the action space is the torque size applied to the pendulum. To  facilitate DQN for choosing the action, we decompose it into 12 parts with equal distances. All the four algorithms do not limit the number of episodes, run 20,000 steps, and repeat 50 rounds. From the results, our algorithm achieves a fast convergence rate and has the optimal average reward with reduced variances.
The right part of Figure ~\ref{fig:exp:perfCurves} compares the performance of the four algorithms under CartPole task. In this task, we need to keep the pole standing, and once it falls, the task is terminated. So we replace the average rewards for each step with the average reward for each episode.
To accelerate convergence, we replace the reward function from the original discrete value of $0/1$ to a continuous function related to observations. The score is higher when the pole is straighter.
All four algorithms run 800 episodes and repeat 10 rounds.
From the results, our algorithm has an excellent average reward while significantly reducing the variance.

%

\begin{figure*}[htb!]
     \begin{center}
    \includegraphics[width=0.4\textwidth]{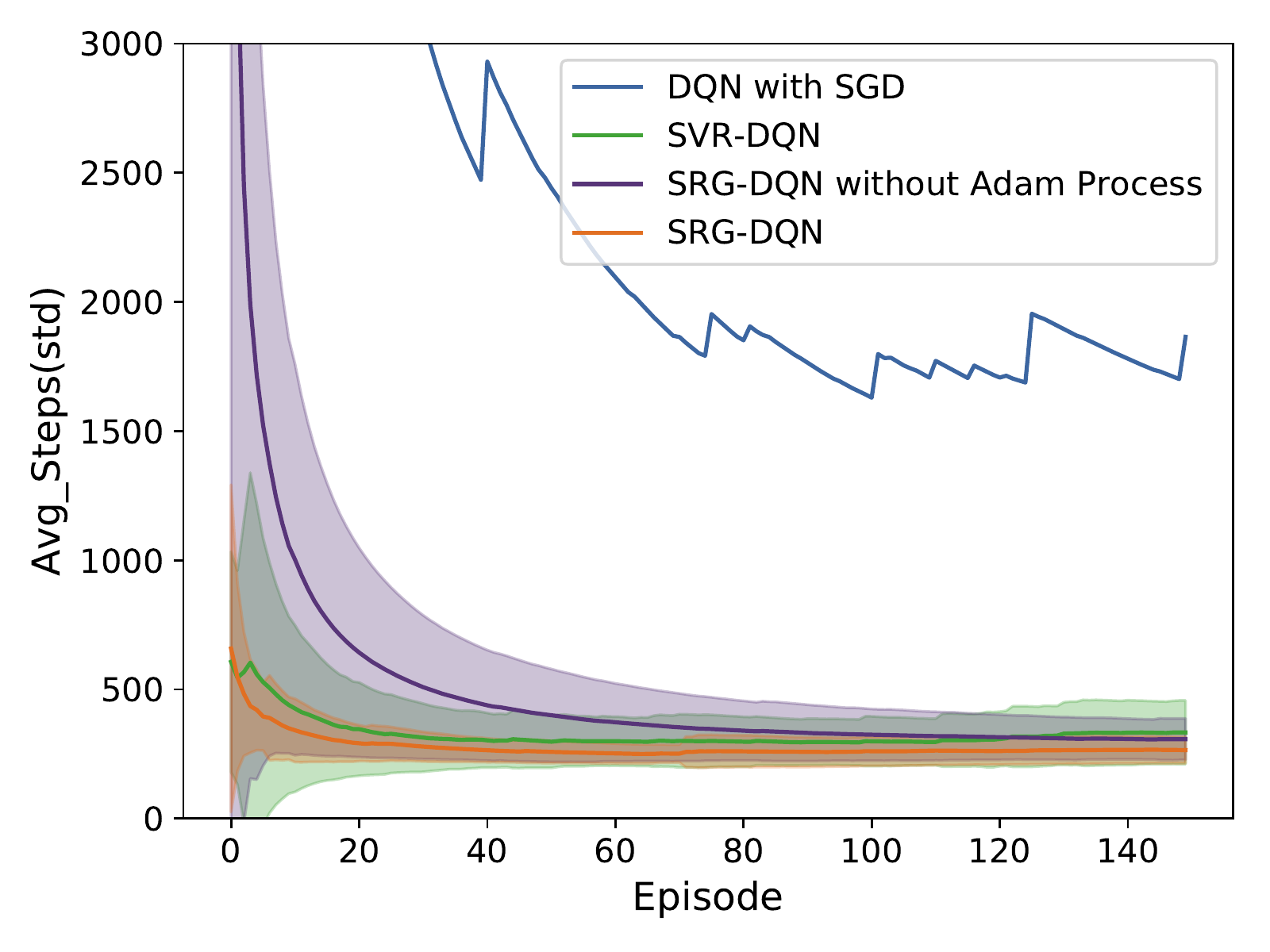}
    \includegraphics[width=0.4\textwidth]{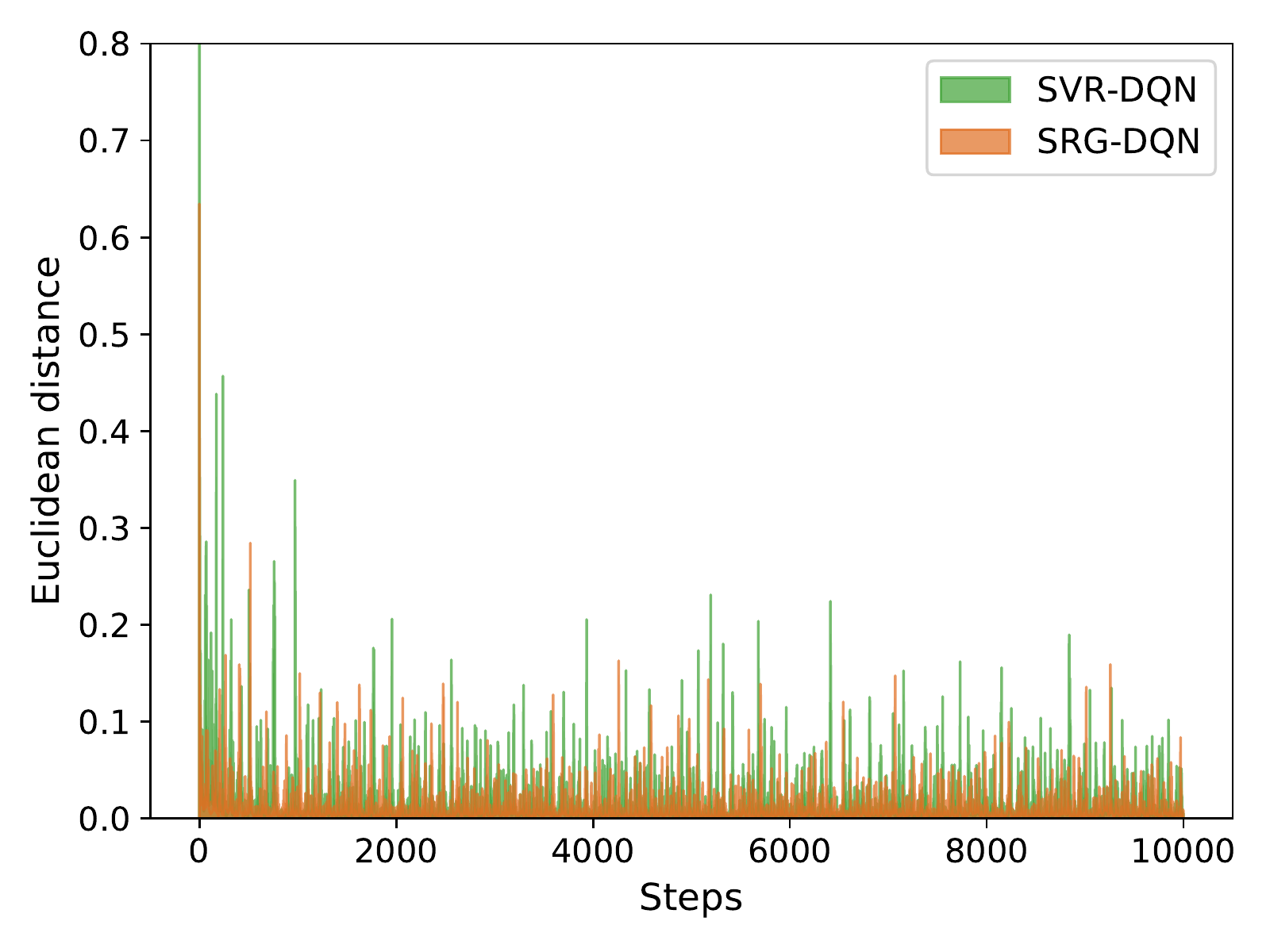}
    \end{center}
    \caption{
    Results of the experimental analysis on `MountainCar-v0' task.
    \emph{Left}: average steps w.r.t. episodes;
    \emph{Right}: $\ell_2$ distances between the exact anchors and the recursive anchors in our SRG-DQN (or the anchors in SVR-DQN).
    The shaded area and the lines have the same meaning as in Figure~\ref{fig:exp:perfCurves}. }
    \label{fig:SRG_DQN:analysis}
   \vskip 0.2in
\end{figure*}

\begin{table*}[!htb]
    \centering
    \footnotesize
    \caption{The comparisons between SVR-DQN and our SRG-DQN
    in terms of the standard deviation of the gradients on `MountainCar-v0'  task,
              where the standard deviation is computed by summing the standard deviations of each element
              in the first-layer network gradient vector.
              We ran the algorithms one hundred thousand steps for the task of the mountain car,
              and recorded the standard deviation once every 1,000 steps for the first 10,000 steps (first three rows)
              and last 10,000 steps (last three rows) .
             }
    \label{table:SRG_DQN:gradient}
    \vskip 0.1in
    \begin{spacing}{1.2}
    \begin{tabular}{l|llllllllll}
    \hline
     \# Steps (first) &1,000 & 2,000 & 3,000 & 4,000 & 5,000 & 6,000 & 7,000 & 8,000 &9,000& 10,000\\
    \hline
            SVR-DQN & 0.229 & 0.790 & 0.430 & \textbf{0.474} & 0.626 & 1.176 & \textbf{0.500} &\textbf{0.388}& 0.638& 0.739\\
            SRG-DQN & \textbf{0.122} & \textbf{0.395} & \textbf{0.389} & 0.520 & \textbf{0.625} & \textbf{0.632} & 0.966 & 0.966& \textbf{0.394}& \textbf{0.562}\\
    \hline
     \# Steps (last) &1,000 & 2,000 & 3,000 & 4,000 & 5,000 & 6,000 & 7,000 & 8,000 &9,000& 10,000\\
    \hline
            SVR-DQN & 0.306 & \textbf{0.286} & 0.293 & \textbf{0.212} & 0.188 & 0.318 & 0.443 & 0.271& 0.193 & 0.205\\
            SRG-DQN & \textbf{0.225} & 0.334 & \textbf{0.254} & 0.353& \textbf{0.154} & \textbf{0.247} & \textbf{0.182} & \textbf{0.229} & \textbf{0.188} & \textbf{0.203}\\
    \hline
    \end{tabular}
    \end{spacing}
\end{table*}
\subsection{Experimental Analysis}
We experimentally analyze the reasons why our SRG-DQN is effective.
We first conducted the episode-average size experiment. In this experiment, four algorithms were run simultaneously with 150 episodes, and 100 rounds were run under the same model parameter settings.
The convergence rate is measured by the average size of the episode, and the stability is evaluated by the standard deviation.
The experimental results are shown in the left part of Figure~\ref{fig:SRG_DQN:analysis}, where the bold line  represents the average episode size of multiple experiments, and the shading represents the standard deviation.
From the results, we can observe that our SRG-DQN has significantly improved the convergence rate and the stability of the agent compared to the traditional DQN with SGD. Compared with the baseline algorithm SVR-DQN, SVR-DQN further shortens the average episode length, reduces the standard deviation. The orange line represents our algorithm SRG-DQN. From the results, Adam process and variance reduction algorithm can play a complementary role, and their combination can further accelerate the algorithm convergence and improve the stability of the agent.
In addition,
we computed the $\ell_2$ distances between the exact anchors and the recursive anchors in our SRG-DQN (or the anchors in SVR-DQN).
From the results about distances in the right part of Figure~\ref{fig:SRG_DQN:analysis},
we can observe that the recursive anchors in our SRG-DQN
can significantly reduce the distances from the exact anchors,
which is another reason why our SRG-DQN can achieve better performances.

To further explore whether our algorithm plays an important role in reducing variance in the process of stochastic optimization,
we compared the performances of our algorithm with SVR-DQN on the variance reduction for gradients.
In these experiments, the algorithms was run 100,000 training steps at the same time, and each ran 50 rounds under the same model parameter settings.
The evaluation criterion in this experiment is the standard deviation of the gradients with respect to the first layer of network parameters.
We calculated the standard deviations of the gradients with respect to the parameters on each dimension separately and then summed them up. 
From the results in Table~\ref{table:SRG_DQN:gradient},
we can observe that,
compared with the existing SVR-DQN, our SRG-DQN significantly reduces the standard deviation, and it is almost completely superior to SVR-DQN at most steps.
Thus, we can conclude that our SRG-DQN converges to the function controlling the variance of the gradients, and achieves an improvement on SVR-DQN for variance reduction, which demonstrates the effectiveness of our stochastic recursive gradient for the variance reduction in DQN.


\section{Conclusion}
This paper proposes a novel deep Q-learning algorithm using stochastic recursive gradients, which reduces the variance of the gradient estimation. The proposed algorithm introduces the recursive framework for updating the stochastic gradient and computing the anchor points. Adam process is involved for achieving a more accurate gradient direction. Theoretical analysis and empirical comparisons showed that the proposed algorithm outperformed the state-of-the-art baselines in terms of reward scores, convergence rate, and stability. The proposed stochastic recursive gradient provides an effective scheme for variance reduction in reinforcement learning.

\section{Appendix A: Proof of the Theorem~\ref{thm:SRG-DQN:IFO:single_outer}}
\begin{proof}[Proof of Theorem~1]
By Lemma~1 in \cite{Nguyen2019Finite},
we have
$$
    \mathbb{E} \left[   F (\theta_{M+1}^{s})   \right] \leq
    \mathbb{E} \left[   F (\theta_{0}^{s}) \right] -
    \dfrac{\eta}{2} \sum_{m=0}^{M} \mathbb{E} \left[  \| \nabla F (\theta_{m}^{s}) \|^2  \right].
$$
Then we get
\begin{align*}
    &\hphantom{{}={}}
    \dfrac{1}{M+1} \sum_{m=0}^{M} \mathbb{E} \left[  \| \nabla F (\theta_{m}^{s}) \|^2  \right] \\
    &\leq
    \dfrac{2}{\eta(M+1)}
        \mathbb{E} \left[   F (\theta_{0}^{s}) -   F (\theta_{M+1}^{s})   \right] \\
    & \leq
    \dfrac{4 B_{\max}}{\eta(M+1)}.
\end{align*}
Setting
$$
    \dfrac{4 B_{\max}}{\eta(M+1)} = \varepsilon,
$$
yielding that
$$
    M = \dfrac{4 B_{\max}}{\eta \varepsilon} - 1.
$$
Since $\eta \leq 2/ [\sqrt{\mu} ( \sqrt{4M +1} +1 ) ]$,
we obtain
$$
    M \geq \dfrac{2 B_{\max} \sqrt{\mu} ( \sqrt{4M +1} +1 ) }{\varepsilon } - 1.
$$
Finally,
we conclude that,
to attain an $\varepsilon$-optimal point requires
$$
    2M \geq \dfrac{4 B_{\max} \sqrt{\mu} ( \sqrt{4M +1} +1 ) }{\varepsilon } - 2
$$
queries to IFO in the inner loop.
\end{proof}

\newpage

\bibliographystyle{aaai}
\bibliography{SRG-DQN}

\begin{thebibliography}{}

\bibitem[\protect\citeauthoryear{Agarwal and Bottou}{2015}]{Agarwal2015Lower}
Agarwal, A., and Bottou, L.
\newblock 2015.
\newblock A lower bound for the optimization of finite sums.
\newblock In {\em Proceedings of the 32nd International Conference on Machine
  Learning},  78--86.

\bibitem[\protect\citeauthoryear{Anschel, Baram, and
  Shimkin}{2017}]{anschel:2017:averaged}
Anschel, O.; Baram, N.; and Shimkin, N.
\newblock 2017.
\newblock Averaged-{DQN}: {V}ariance reduction and stabilization for deep
  reinforcement learning.
\newblock In {\em Proceedings of the 34th International Conference on Machine
  Learning},  176--185.

\bibitem[\protect\citeauthoryear{Defazio, Bach, and
  Lacoste-Julien}{2014}]{defazio:2014:saga}
Defazio, A.; Bach, F.; and Lacoste-Julien, S.
\newblock 2014.
\newblock {SAGA}: {A} fast incremental gradient method with support for
  non-strongly convex composite objectives.
\newblock In {\em Advances in Neural Information Processing Systems 27},
  1646--1654.

\bibitem[\protect\citeauthoryear{Du \bgroup et al\mbox.\egroup
  }{2017}]{Du2017Stochastic}
Du, S.~S.; Chen, J.; Li, L.; Xiao, L.; and Zhou, D.
\newblock 2017.
\newblock Stochastic variance reduction methods for policy evaluation.
\newblock In {\em Proceedings of the 34th International Conference on Machine
  Learning},  1049--1058.

\bibitem[\protect\citeauthoryear{Fortunato \bgroup et al\mbox.\egroup
  }{2017}]{fortunato:2017:noisy}
Fortunato, M.; Azar, M.~G.; Piot, B.; Menick, J.; Osband, I.; Graves, A.; Mnih,
  V.; Munos, R.; Hassabis, D.; Pietquin, O.; et~al.
\newblock 2017.
\newblock Noisy networks for exploration.
\newblock {\em arXiv preprint arXiv:1706.10295}.

\bibitem[\protect\citeauthoryear{Hessel \bgroup et al\mbox.\egroup
  }{2018}]{hessel:2018:rainbow}
Hessel, M.; Modayil, J.; Van~Hasselt, H.; Schaul, T.; Ostrovski, G.; Dabney,
  W.; Horgan, D.; Piot, B.; Azar, M.; and Silver, D.
\newblock 2018.
\newblock Rainbow: {C}ombining improvements in deep reinforcement learning.
\newblock In {\em Proceedings of the 32rd AAAI Conference on Artificial
  Intelligence},  3215--3222.

\bibitem[\protect\citeauthoryear{Johnson and
  Zhang}{2013}]{johnson:2013:accelerating}
Johnson, R., and Zhang, T.
\newblock 2013.
\newblock Accelerating stochastic gradient descent using predictive variance
  reduction.
\newblock In {\em Advances in Neural Information Processing Systems 26},
  315--323.

\bibitem[\protect\citeauthoryear{Kingma and Ba}{2014}]{Kingma2014Adam}
Kingma, D., and Ba, J.
\newblock 2014.
\newblock Adam: {A} method for stochastic optimization.
\newblock {\em arXiv preprint arXiv:1412.6980}.

\bibitem[\protect\citeauthoryear{Li, Ma, and
  Giannakis}{2019}]{Li2019Convergence}
Li, B.; Ma, M.; and Giannakis, G.~B.
\newblock 2019.
\newblock On the convergence of {SARAH} and beyond.
\newblock {\em arXiv:1906.02351}.

\bibitem[\protect\citeauthoryear{Mnih \bgroup et al\mbox.\egroup
  }{2013}]{Mnih:2013:DQN}
Mnih, V.; Kavukcuoglu, K.; Silver, D.; Graves, A.; Antonoglou, I.; Wierstra,
  D.; and Riedmiller, M.
\newblock 2013.
\newblock Playing atari with deep reinforcement learning.
\newblock {\em arXiv preprint arXiv:1312.5602}.

\bibitem[\protect\citeauthoryear{Mnih \bgroup et al\mbox.\egroup
  }{2015}]{Mnih:2015:DQN}
Mnih, V.; Kavukcuoglu, K.; Silver, D.; Rusu, A.~A.; Veness, J.; Bellemare,
  M.~G.; Graves, A.; Riedmiller, M.; Fidjeland, A.~K.; Ostrovski, G.; Petersen,
  S.; Beattie, C.; Sadik, A.; Antonoglou, I.; King, H.; Kumaran, D.; Wierstra,
  D.; Legg, S.; and Hassabis, D.
\newblock 2015.
\newblock Human-level control through deep reinforcement learning.
\newblock {\em Nature} 518(7540):529--533.

\bibitem[\protect\citeauthoryear{Nguyen \bgroup et al\mbox.\egroup
  }{2017}]{nguyen:2017:sarah}
Nguyen, L.~M.; Liu, J.; Scheinberg, K.; and Tak{\'a}{\v{c}}, M.
\newblock 2017.
\newblock {SARAH}: {A} novel method for machine learning problems using
  stochastic recursive gradient.
\newblock In {\em Proceedings of the 34th International Conference on Machine
  Learning},  2613--2621.

\bibitem[\protect\citeauthoryear{Nguyen \bgroup et al\mbox.\egroup
  }{2019}]{Nguyen2019Finite}
Nguyen, L.~M.; van Dijk, M.; Phan, D.~T.; and Nguyen, P.~H.
\newblock 2019.
\newblock Finite-sum smooth optimization with {SARAH}.
\newblock {\em arXiv:1901.07648v2}.

\bibitem[\protect\citeauthoryear{Papini \bgroup et al\mbox.\egroup
  }{2018}]{Papini2018Stochastic}
Papini, M.; Binaghi, D.; Canonaco, G.; Pirotta, M.; and Restelli, M.
\newblock 2018.
\newblock Stochastic variance-reduced policy gradient.
\newblock In {\em Proceedings of the 35th International Conference on Machine
  Learning},  4023--4032.

\bibitem[\protect\citeauthoryear{Romoff \bgroup et al\mbox.\egroup
  }{2018}]{romoff:2018:reward}
Romoff, J.; Henderson, P.; Pich{\'e}, A.; Francois-Lavet, V.; and Pineau, J.
\newblock 2018.
\newblock Reward estimation for variance reduction in deep reinforcement
  learning.
\newblock {\em arXiv preprint arXiv:1805.03359}.

\bibitem[\protect\citeauthoryear{Roux, Schmidt, and Bach}{2012}]{roux:2012:sag}
Roux, N.~L.; Schmidt, M.; and Bach, F.~R.
\newblock 2012.
\newblock A stochastic gradient method with an exponential convergence rate for
  finite training sets.
\newblock In {\em Advances in Neural Information Processing Systems 25},
  2663--2671.

\bibitem[\protect\citeauthoryear{Sabry and Khalifa}{2019}]{Sabry2019Reduction}
Sabry, M., and Khalifa, A. M.~A.
\newblock 2019.
\newblock On the reduction of variance and overestimation of deep {Q}-learning.
\newblock {\em arXiv preprint arXiv:1910.05983}.

\bibitem[\protect\citeauthoryear{Schaul \bgroup et al\mbox.\egroup
  }{2015}]{schaul:2015:prioritized}
Schaul, T.; Quan, J.; Antonoglou, I.; and Silver, D.
\newblock 2015.
\newblock Prioritized experience replay.
\newblock {\em arXiv preprint arXiv:1511.05952}.

\bibitem[\protect\citeauthoryear{van Hasselt, Guez, and
  Silver}{2016}]{van:2016:DDQN}
van Hasselt, H.; Guez, A.; and Silver, D.
\newblock 2016.
\newblock Deep reinforcement learning with double {Q}-learning.
\newblock In {\em Proceedings of the 30th AAAI Conference on Artificial
  Intelligence},  2094--2100.

\bibitem[\protect\citeauthoryear{Wang \bgroup et al\mbox.\egroup
  }{2015}]{wang:2015:dueling}
Wang, Z.; Schaul, T.; Hessel, M.; Van~Hasselt, H.; Lanctot, M.; and De~Freitas,
  N.
\newblock 2015.
\newblock Dueling network architectures for deep reinforcement learning.
\newblock {\em arXiv preprint arXiv:1511.06581}.

\bibitem[\protect\citeauthoryear{Xu, Liu, and Peng}{2017}]{Xu2017Stochastic}
Xu, T.; Liu, Q.; and Peng, J.
\newblock 2017.
\newblock Stochastic variance reduction for policy gradient estimation.
\newblock {\em arXiv:1710.06034}.

\bibitem[\protect\citeauthoryear{Zhao and Peng}{2019}]{zhao:2019:stochastic}
Zhao, W.-Y., and Peng, J.
\newblock 2019.
\newblock Stochastic variance reduction for deep {Q}-learning.
\newblock In {\em Proceedings of the 18th International Conference on
  Autonomous Agents and MultiAgent Systems},  2318--2320.

\bibitem[\protect\citeauthoryear{Zhao \bgroup et al\mbox.\egroup
  }{2019}]{zhao:2019:stochastic-arXiv}
Zhao, W.; Guan, X.; Liu, Y.; Zhao, X.; and Peng, J.
\newblock 2019.
\newblock Stochastic variance reduction for deep {Q}-learning.
\newblock {\em arXiv preprint arXiv:1905.08152}.

\end{thebibliography}

\end{document}